\documentclass[conference]{IEEEtran}
\IEEEoverridecommandlockouts

\newif\ifextended
\extendedfalse

\usepackage{cite}
\usepackage{amsmath,amssymb}
\usepackage{graphicx}
\usepackage{booktabs}
\usepackage{siunitx}
\usepackage{xcolor}
\usepackage{tikz}
\usetikzlibrary{positioning,arrows.meta}

\newif\ifsubmissionversion
\submissionversiontrue          

\ifsubmissionversion

  \newcommand{\raddar}[1]{#1}
  \newcommand{\raddleo}[1]{#1}
  \newcommand{\raddraj}[1]{#1}
  \newcommand{\rdelar}[1]{}
  \newcommand{\rdelleo}[1]{}
  \newcommand{\rdelraj}[1]{}

\else

  \definecolor{editar}{rgb} {0.00,0.30,0.75}   
  \definecolor{editleo}{rgb}{0.00,0.45,0.35}   
  \definecolor{editraj}{rgb}{0.50,0.15,0.55}   
  \definecolor{editcut}{gray}{0.55}            

  \newcommand{\raddar}[1]{{\color{editar}#1}}
  \newcommand{\raddleo}[1]{{\color{editleo}#1}}
  \newcommand{\raddraj}[1]{{\color{editraj}#1}}
  \newcommand{\rdelar}[1]{{\color{editcut}[cut: #1]}}
  \newcommand{\rdelleo}[1]{{\color{editcut}[cut: #1]}}
  \newcommand{\rdelraj}[1]{{\color{editcut}[cut: #1]}}

\fi

\usepackage[colorlinks=true,allcolors=black]{hyperref}

\newif\ifdraftnotes
\draftnotestrue

\newcommand{\armOurs}{64.4\%}
\newcommand{\armLift}{63.6\%}
\newcommand{\armPlat}{51.1\%}
\newcommand{\armNaive}{54.4\%}
\newcommand{\armMem}{68.3\%}
\newcommand{\dLift}{0.8}          
\newcommand{\tLift}{0.23}
\newcommand{\dPlat}{13.3}         
\newcommand{\tPlat}{2.77}
\newcommand{\dNaive}{10.0}        
\newcommand{\tNaive}{2.38}
\newcommand{\dMem}{3.9}           
\newcommand{\tMem}{1.13}
\newcommand{\ndtwOurs}{0.360}
\newcommand{\ndtwLift}{0.279}
\newcommand{\ndtwPlat}{0.352}
\newcommand{\ndtwNaive}{0.337}
\newcommand{\ndtwMem}{0.394}
\newcommand{\tlrOurs}{1.23}
\newcommand{\tlrLift}{1.53}
\newcommand{\tlrPlat}{0.79}
\newcommand{\tlrNaive}{1.00}
\newcommand{\tlrMem}{1.20}
\newcommand{\tlrRef}{\SI{11.0}{\meter}}    

\newcommand{\apprOurs}{\SI{1.58}{\meter}}
\newcommand{\apprLift}{\SI{1.73}{\meter}}
\newcommand{\apprPlat}{\SI{2.10}{\meter}}

\newcommand{\dApprPlat}{\SI{0.52}{\meter}}   
\newcommand{\tApprPlat}{3.42}
\newcommand{\signApprPlat}{24 of 30}
\newcommand{\dApprNaive}{\SI{0.27}{\meter}}  
\newcommand{\tApprNaive}{2.25}

\newcommand{\estModel}{\SI{2.80}{\meter}}   
\newcommand{\estLift}{\SI{0.16}{\meter}}    

\newcommand{\aimGapMed}{\SI{1.47}{\meter}}     
\newcommand{\settleGapMed}{\SI{0.15}{\meter}}  

\newcommand{\dSteps}{1.45}                     
\newcommand{\tSteps}{2.86}

\newcommand{\repeatOurs}{63.5\% and 62.2\%}   
\newcommand{\repeatLift}{57.7\% and 63.5\%}   
\newcommand{\repeatWorst}{5.8}                
\newcommand{\sdWithin}{1.07}                  
\newcommand{\sdBetween}{1.84}                 
\newcommand{\minDetect}{9.4}                  

\newcommand{\nQuestions}{30}
\newcommand{\nQuestionsOld}{26}  
\newcommand{\nScenes}{15}

\newcommand{\tol}{\SI{1.25}{\meter}}

\newcommand{\ttQuestions}{45}
\newcommand{\ttScenes}{15}
\newcommand{\ttIoUThresh}{0.1}         
\newcommand{\ttFullHits}{10}
\newcommand{\ttFullPct}{22\%}          
\newcommand{\ttFullIoU}{0.060}         

\newcommand{\ttNaivePct}{18\%}         
\newcommand{\ttNaiveIoU}{0.054}

\newcommand{\ttModelHits}{0}
\newcommand{\ttModelPct}{0\%}          
\newcommand{\ttModelIoU}{0.000}

\newcommand{\ttPriorSize}{100}         
\newcommand{\ttGraphBaseline}{52.4}    
\newcommand{\ttGraphNaive}{43.8}       
\newcommand{\ttGraphBigger}{30}        
\newcommand{\ttCaseBaseDist}{\SI{0.23}{\meter}}   
\newcommand{\ttCaseMetricEst}{\SI{3.5}{\meter}}   
\newcommand{\ttCaseMetricDist}{\SI{6.0}{\meter}}  
\newcommand{\ttCaseFullObjs}{25}       
\newcommand{\ttCaseNaiveObjs}{10}      

\newcommand{\percHz}{2}              
\newcommand{\detInliers}{10}         
\newcommand{\detNovelView}{0.3}      
\newcommand{\detScore}{0.4}          
\newcommand{\detNMS}{0.5}            
\newcommand{\detSAM}{0.8}            
\newcommand{\qBudgetMin}{10}         
\newcommand{\answerReserveSec}{60}   
\newcommand{\navCutoffSec}{540}      

\begin{document}

\title{AnchorVLN: Geometry-Anchored Vision-Language
  Grounding Reasoning for Open-Vocabulary Navigation}

\author{\IEEEauthorblockN{Long Giang Vu\textsuperscript{*}, Chengkai Yao\textsuperscript{*}, Yuxin Liu, FNU Aryan, Rajath Chandrashekar Aralikatti}
\IEEEauthorblockA{
giangvl.cs@gmail.com, chengkaiyao825@gmail.com, yuxin4500@gmail.com, \\aryanmangal2005@gmail.com, rajathcaralikatti@gmail.com%
\thanks{\textsuperscript{*}These authors contributed equally.}}}

\maketitle

\begin{abstract}
\raddar{Vision-Language Navigation (VLN) in previously unseen indoor
environments is useful in real-world robotics, where an agent
must follow natural-language instructions, locate referred
objects, and answer spatial questions without a pre-built map
or a fixed object vocabulary. Recently, multimodal
vision-language models (VLMs) have shown strong open-vocabulary
grounding and zero-shot reasoning, making them a natural
semantic front-end for such agents. One common problem is these
VLMs' inability to emit reliable metric quantities such as
range, bearing, and comparative spatial relations directly
from images. Existing approaches address this by folding
geometry into a hand-engineered pipeline, or by asking the
model to output waypoints; both require rewriting the control
stack for each new robot, task, or vocabulary, which does not
scale.
This work designs \textbf{AnchorVLN}, an open-vocabulary VLN
system built on a single rule (\emph{the VLM proposes
semantics; geometry decides metrics}), realised as
\textsc{Embodied-Nav-MCP}, a Model Context Protocol (MCP) server
that a VLM agent drives at inference time through a compact set
of callable tools. Because no tool accepts a distance in metres
or a bearing in radians, the schema itself enforces the
boundary between semantics and geometry without rewriting the
downstream autonomy stack. We benchmark the server on both
tasks of the CMU Vision-Language Navigation Challenge 2026: all
\nQuestions{} instruction-following questions over \nScenes{}
scenes, and a frozen \ttQuestions{}-question object-reference
set on the same scenes. \raddraj{The full system achieves
64.4\% on instruction following, dropping by 13.3 percentage
points without controller modeling ($t=2.77$, paired over
questions).} \raddraj{On object reference, geometric anchoring
clears the challenge's overlap threshold on 10 of 45 questions
(versus 0 of 45 for direct model coordinate estimation),
reducing median center error from 3.37~m to 2.48~m.}}
\end{abstract}


%

\section{Introduction}
\label{sec:intro}
\begin{figure}[t]
  \centering
  \includegraphics[width=\columnwidth]{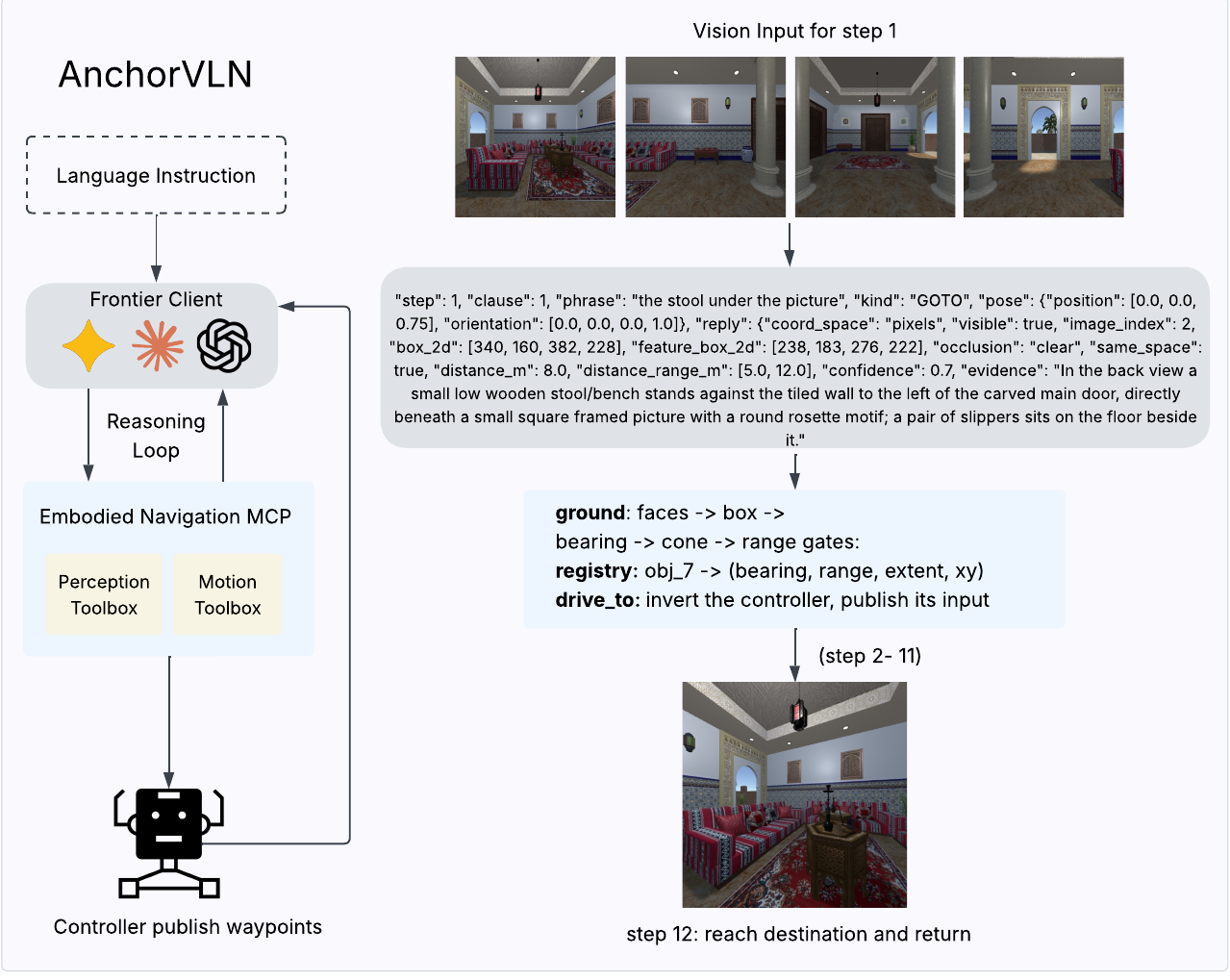}
  \caption{Three authorities, and the two boundaries between them. The client
  says \emph{which}, the server decides \emph{how far} or declines to, and the
  controller decides where the vehicle comes to rest. Only phrases and handles
  cross the upper boundary, so a coordinate the client invented cannot reach
  the robot. The lower boundary lies wherever the policy stops owning
  actuation; the server models the controller below it rather than commanding
  it, and porting to another platform replaces that model alone.}
  \label{fig:system}
\end{figure}

Developing robots that can navigate previously unseen indoor environments
is a longstanding problem in robotics. Early systems solved it
geometrically: lidar odometry and visual SLAM~\cite{loam,orbslam2}
together with closed-vocabulary detectors~\cite{maskrcnn} built a
metric map, and a hand-engineered planner queried that map for goals.
The maps are consistent, but the object vocabulary is frozen before
deployment, so free-form referring expressions and compositional
phrases are out of reach without extra supervision.
Vision-and-Language Navigation (VLN) recast the same problem as
following natural-language instructions in unseen
buildings~\cite{vln,vlnce}. In the last few years that line of work
has moved from specialist policies trained on demonstrations to
agents that put a multimodal vision-language model (VLM) in the
loop, using the model as an open-vocabulary semantic
front-end~\cite{navgpt2,navid,vlmaps,navila}.

These VLMs are good at what language-and-vision pretraining
teaches: open-vocabulary recognition, parsing referring expressions,
and multi-step reasoning over an instruction. Recent VLN systems
use that strength in a few patterns---a VLM that emits the next
action from video~\cite{navid,navila}, a modular object-goal stack
that reserves the model for high-level choices~\cite{sysnav}, or a
planner that converts image evidence to metres with camera geometry
so the model never outputs 3D coordinates~\cite{aoplanner}. The
failure that remains is metric. The same model that grounds
\emph{the lantern closest to the fan decoration} will still invent
a range or pick the wrong comparative when asked to decide in
metres, and the usual remedies---rewriting the perception-and-control
pipeline for each robot, or leaving waypoints to the model---do not
scale. We therefore treat the failure as an interface problem:
\emph{the VLM proposes semantics; geometry decides metrics.}

We introduce {AnchorVLN} as shown in Fig.~\ref{fig:system}, an open-vocabulary VLN system
built on that rule and realised as \textsc{Embodied-Nav-MCP}, a
Model Context Protocol (MCP)~\cite{mcp} server that a VLM agent
drives at inference time through a compact set of tools
(Sec.~\ref{sec:method}). \rdelraj{Only phrases and opaque handles cross the
client--server boundary: \texttt{ground} lifts a phrase through the
onboard lidar, comparatives are tools over handles, and
\texttt{drive\_to} publishes a request to an autonomy stack we do
not own and cannot modify} \raddraj{Communication across the client--server boundary is strictly restricted to text phrases and opaque handles (e.g., \texttt{obj\_1}): \texttt{ground} detects referents and extracts metric depth via onboard LiDAR, comparative tools evaluate relative spatial queries over handles, and \texttt{drive\_to} dispatches waypoints to an unmodifiable autonomy stack}~\cite{aedev,tare,farplanner}. Because no
tool accepts a distance in metres or a bearing in radians, \rdelraj{a number
the model invented} \raddraj{model-hallucinated coordinates} cannot reach the robot. Our contributions are:
\begin{itemize}
  \item A \emph{semantic--geometric decoupling} enforced by the MCP
        schema: no tool takes a metric argument, so the client does
        language-to-pixel grounding and lidar produces the metres.
  \item \textsc{Embodied-Nav-MCP}\footnote{\url{https://github.com/aryanmangal769/embodied-nav-mcp}},
        a reasoning-driven tool server that a VLM agent composes at
        inference time, replacing hand-scripted per-tick planners.
  \item Evaluation on the CMU VLN Challenge 2026 development set:
        \armOurs{} on instruction following (\dPlat{}-point drop
        without the controller model) and \ttFullPct{} vs.\
        \ttModelPct{} 3D-IoU hits when geometry, not the model,
        emits object coordinates.
\end{itemize}

\section{Related Work}
\label{sec:related}
\subsection{\raddar{Vision-Language Navigation}}

\raddar{Indoor navigation was first a geometric problem: lidar odometry and visual
SLAM~\mbox{\cite{loam,orbslam2,airloc,gradlidar,pypose}} with closed-vocabulary detectors~\mbox{\cite{maskrcnn}}
produce a metric map, and a planner queries it for goals. The stack this
challenge ships is in that lineage---exploration, route planning, and
waypoint following on a vehicle the policy does not
own~\mbox{\cite{aedev,tare,farplanner}}. Vision-and-Language Navigation
(VLN)~\mbox{\cite{vln}} recast the same setting as following natural-language
instructions on a discrete nav-graph; VLN-CE~\mbox{\cite{vlnce}} lifted it to
continuous control. Parallel object-goal lines study language-driven search
without a full instruction, including zero-shot
ObjectNav~\mbox{\cite{cow,esc}}. Referential 3D benchmarks then made the
target an instance, not a class: VLA-3D~\mbox{\cite{vla3d}} supplies the
annotations this challenge uses, and IRef-VLA~\mbox{\cite{irefvla}} studies
iterative 3D referring. Closest to our platform, SysNav~\mbox{\cite{sysnav}} is a
three-level ObjectNav system from the same laboratory: it decouples
semantic reasoning, navigation planning and motion control, and reserves
the VLM for room-level decisions because a model is too slow to sit in the
control loop---a decoupling we adopt and say is theirs.
We differ in that we do not own the controller, and we score ordered
trajectories with object-level referring expressions rather than ObjectNav
arrival.}

\subsection{\raddar{Vision-Language Models for VLN}}

\raddar{Multimodal models changed who does the language. LM-Nav~\mbox{\cite{lmnav}}
parses landmarks with an LLM and grounds them with CLIP on a pre-built
topological map; NavGPT~\mbox{\cite{navgpt}} and
NavGPT-2~\mbox{\cite{navgpt2}} use an LLM as an explicit navigational reasoner.
Video VLMs then predicted the next action from a monocular
stream~\mbox{\cite{navid,streamvln}}, and dual-system designs pair a slow VLM
planner with a fast policy the authors
train~\mbox{\cite{navila,groundslow}}. On a real robot, Xu~\emph{et
al.}~\mbox{\cite{xu-vln}} parse instructions with an LLM, build an online
visual-language map, and hand waypoints to a learned local controller.
AO-Planner~\mbox{\cite{aoplanner}} is the nearest neighbour to our metric
rule: it refuses to ask the model for 3D coordinates and converts image
evidence to metres with camera parameters and depth, but it grounds
traversable ground in a space it owns. Open-vocabulary maps---VLMaps~\mbox{\cite{vlmaps}},
ConceptFusion~\mbox{\cite{conceptfusion}}, OpenScene~\mbox{\cite{openscene}},
CLIP-Fields~\mbox{\cite{clipfields}}, NLMap~\mbox{\cite{nlmap}}---fuse language
features into a reconstruction that can be queried after the fact; we
defer vocabulary to query time as a design choice, not a result. Xia~\emph{et
al.}~\mbox{\cite{bottlenecks}} argue from benchmarks that 3D perception
accuracy saturates; we agree that recognition saturates, and treat ranging
as a separate, confident failure.
We keep the VLM on language-to-pixel grounding and take metres from
lidar, on a controller we cannot change.}

\subsection{\raddar{Agentic Systems for Real-World Tasks}}

\raddar{Outside navigation, large models have been used as agents that call tools
and skills rather than outputting motor commands. ReAct~\mbox{\cite{react}}
interleaves reasoning traces with actions over APIs; Toolformer~\mbox{\cite{toolformer}}
shows that language models can learn when to invoke those APIs.
On physical robots, SayCan~\mbox{\cite{saycan}} grounds an LLM in pretrained
skill affordances on a mobile manipulator, Inner Monologue~\mbox{\cite{innermonologue}}
closes the loop with language feedback from the scene, and
PaLM-E~\mbox{\cite{palme}} injects multimodal observations into the same
planner. The Model Context Protocol~\mbox{\cite{mcp}} standardises that
pattern as a client--server tool schema.
We use that pattern; the difference is that every argument is a phrase
or a handle, not a number.}

%
%
%
%
%
\section{Methodology}
This section presents the methodology of our work. We first describe the interface, the tools the VLM client calls and the phrases and handles that cross that boundary. We then describe perception, how phrases are grounded, comparatives are resolved, and an object graph is kept. We close with motion, how the server drives a controller it does not own.
\label{sec:method}
\begin{table}[t]
  \centering
  \caption{The \textsc{Embodied-Nav-MCP} tool interface. Every input is a
  phrase or a handle from an earlier call --- \textbf{no input is a number} ---
  so the client can name a measurement the system has made but cannot author
  one.}
  \label{tab:tools}
  \footnotesize
  \setlength{\tabcolsep}{4pt}
  \begin{tabular}{@{}p{2.45cm}p{2.15cm}p{3.15cm}@{}}
    \toprule
    Tool & Input & Returns \\
    \midrule
    \multicolumn{3}{@{}l}{\emph{Perception --- query or steer; the robot stays put}} \\
    \texttt{ground}          & phrase           & handles, or a refusal \\
    \texttt{expand\_vocab}   & phrase           & classes added \\
    \texttt{verify\_present} & phrase           & handle, or not found \\
    \texttt{nearest}         & handles, anchor  & handle \\
    \texttt{farthest}        & handles, anchor  & handle \\
    \texttt{between}         & two handles      & handle (a passage) \\
    \addlinespace[3pt]
    \multicolumn{3}{@{}l}{\emph{Motion --- the vehicle moves}} \\
    \texttt{drive\_to}       & handle           & arrived, moved, stalled \\
    \texttt{explore}         & handles to avoid & handles newly seen \\
    \texttt{orbit}           & handle           & same handle, refined \\
    \addlinespace[3pt]
    \multicolumn{3}{@{}l}{\emph{Output}} \\
    \texttt{publish}         & handle           & --- \\
    \bottomrule
  \end{tabular}
\end{table}

\subsection{The interface}
\label{sec:method-interface}
\rdelar{Our \textsc{Embodied-Nav-MCP} exposes the perception and navigation stack as
the 10 tools of Table~\mbox{\ref{tab:tools}}. Every argument they take is a phrase or a \emph{handle} --- an
opaque name such as \texttt{obj\_7} for a record the server holds, carrying a
bearing, a range, an extent and a map-frame coordinate. A handle is a name and
not a value: nothing the client can do turns one into a location, and the
server resolves only identifiers it has itself issued. The VLM is called in
three places: (1) as the MCP client, which reads the request in natural
language and decides which tool to call next; (2) inside \texttt{ground}, the
one tool that calls it directly, where it is given 2D images and an object
name as the target; and
(3) inside the perception process that maintains the \emph{object graph},
one record per object the robot has seen --- the records handles name.
Another 2 tools reach that process without calling the VLM themselves ---
\texttt{expand\_vocab} sets the vocabulary it searches for, and
\texttt{verify\_present} lowers the threshold at which it will admit one
class. The remaining 7 tools have no VLM in them: comparing two positions,
choosing a waypoint, circling a reached object and picking a
frontier are arithmetic over records the graph already holds.}

\rdelar{An instruction is decomposed once into ordered clauses: destinations,
keep-outs, and \emph{passages} --- somewhere to go \emph{through} rather than
\emph{to}, as \emph{between the TV and the bed} is satisfied by crossing
rather than by arriving. Working through them in order is what lets a late
\emph{comparative} --- a phrase that picks its target out by its relation to
another object, as in \emph{the chair closest to the TV} --- refer to an
\emph{anchor} that an earlier clause established. The anchor is the object
measured from, and never the answer.}

\rdelar{Below the interface the server speaks \textsc{ROS} to the platform, and that
boundary is narrow in both directions. It reads four topics --- the panoramic
camera, the registered LiDAR scan, the terrain map and odometry --- and writes
one: a single \texttt{Pose2D} waypoint. \texttt{ground} consumes the first
two, \texttt{drive\_to} writes the last, and a call such as
\texttt{drive\_to(obj\_7)} leaves the server as one pose on
\texttt{/way\_point\_with\_heading} and nothing else. Arrival is read back off
odometry rather than reported by the platform. \texttt{publish} is the one
tool whose output leaves this loop instead of steering it: it emits a handle
the client already holds, so the box the system answers with is geometry the
graph committed rather than a coordinate a model authored.}

\raddar{In this part we describe the interface: how the VLM client talks to the
stack, and what is allowed to cross that boundary. \textsc{Embodied-Nav-MCP}
exposes the perception and navigation stack as the 10 tools of
Table~\mbox{\ref{tab:tools}}. Every argument they take is a phrase or a \emph{handle}, an
opaque name such as \texttt{obj\_7} for a record the server holds, carrying a
bearing, a range, an extent and a map-frame coordinate. A handle is a name and
not a value: nothing the client can do turns one into a location, and the
server resolves only identifiers it has itself issued. The VLM is called in
three places: (1) as the MCP client, which reads the request in natural
language and decides which tool to call next; (2) inside \texttt{ground}, the
one tool that calls it directly, where it is given 2D images and an object
name as the target; and (3) inside the perception process that maintains the
\emph{object graph}, one record per object the robot has seen, the records
handles name. \texttt{expand\_vocab} and \texttt{verify\_present} reach that
process without calling the VLM; the remaining tools have no VLM in them, and
are arithmetic over records the graph already holds.}

\raddar{An instruction is decomposed once into ordered clauses: destinations,
keep-outs, and \emph{passages}, somewhere to go \emph{through} rather than
\emph{to}, as \emph{between the TV and the bed} is satisfied by crossing
rather than by arriving. Working through them in order is what lets a late
\emph{comparative}, a phrase that picks its target out by its relation to
another object, as in \emph{the chair closest to the TV}, refer to an
\emph{anchor} that an earlier clause established.  Below the interface the server speaks
\textsc{ROS} to the platform: it reads the panoramic camera, the registered
LiDAR scan, the terrain map and odometry, and writes a single \texttt{Pose2D}
waypoint. \texttt{ground} consumes the camera and scan, \texttt{drive\_to}
writes the waypoint, arrival is read back off odometry rather than reported
by the platform, and \texttt{publish} is the one tool whose output leaves
this loop, so the box the system answers with is geometry the graph committed
rather than a coordinate a model authored.}

\subsection{Perception}

\rdelar{These 6 tools are how the client learns what is in the scene. None of them
moves the robot.}

\rdelar{\texttt{ground} is the load-bearing call. It unwraps the onboard panorama into
a few overlapping gnomonic faces, which keeps resolution on small distant
objects; the VLM returns a box per candidate; the box centre is unprojected to
a bearing; and the \emph{median} range of scan returns in a narrow cone about
that bearing becomes depth, median so that foreground and background surfaces
in one cone cannot average into a plausible wrong number. What comes back is
handles, or a refusal.}

\rdelar{Refusal is a feature, because a confident wrong coordinate is driven to at
speed. Three gates abort the lift, each a check between two measurements no
model sees: an \emph{elevation floor} rejects a bearing below the scanner's
lowest ray, where the cone samples floor; an \emph{implied-size} gate rejects
a box whose extent at the returned range contradicts the noun; an
\emph{empty-cone} test rejects a bearing with too few returns to cluster. A
gated lift still yields a handle, marked unmeasured, and \texttt{drive\_to} on
it steps a bounded distance along the known bearing --- often enough, since
moving changes the geometry that caused the refusal.}

\rdelar{\texttt{nearest}, \texttt{farthest} and \texttt{between} resolve comparatives,
and an instruction needs them whenever it identifies its target through an
anchor. The answer is arithmetic over records the graph already holds, and it
is a tool rather than a client judgement because the client is bad at it:
asked which candidate is closest to an anchor, a model reliably returns the
\emph{anchor}; asked only to enumerate the candidates and the anchor, it never
gets the chance. Because the operands come from the graph rather than the
current image, a comparative ranges over every instance the robot has ever
seen. \texttt{between} is the exception in what it returns --- not an object
but a passage: the accessible midpoint of two handles.}

\rdelar{\texttt{expand\_vocab} and \texttt{verify\_present} maintain the vocabulary
the perception process searches for, and through it what the object graph
comes to hold. That graph lives in the server: following
\textsc{SysNav}~\mbox{\cite{sysnav}}, each LiDAR sweep and the objects recognised in
it are folded into persistent records keyed by identity, so an object seen
again from a new viewpoint updates one record rather than spawning a second,
and its extent tightens as occluded faces come into view. It is where every
handle resolves, and where object reference reads the 3D box it answers with.}

\raddar{In this part we describe perception: the tools that tell the client what is
in the scene, none of which moves the robot. \texttt{ground} is the
load-bearing call. It unwraps the onboard panorama into a few overlapping
gnomonic faces, which keeps resolution on small distant objects; the VLM
returns a box per candidate; the box centre is unprojected to a bearing; and
the \emph{median} range of scan returns in a narrow cone about that bearing
becomes depth, median so that foreground and background surfaces in one cone
cannot average into a plausible wrong number. What comes back is handles, or
a refusal. Refusal is a feature, because a confident wrong coordinate is
driven to at speed. Three gates abort the lift, each a check between two
measurements no model sees: an \emph{elevation floor} rejects a bearing below
the scanner's lowest ray, where the cone samples floor; an \emph{implied-size}
gate rejects a box whose extent at the returned range contradicts the noun;
an \emph{empty-cone} test rejects a bearing with too few returns to cluster.
A gated lift still yields a handle, marked unmeasured, and \texttt{drive\_to}
on it steps a bounded distance along the known bearing, often enough, since
moving changes the geometry that caused the refusal.}

\raddar{\texttt{nearest}, \texttt{farthest} and \texttt{between} resolve comparatives.
The answer is arithmetic over records the graph already holds, and it is a
tool rather than a client judgement because the client is bad at it: asked
which candidate is closest to an anchor, a model reliably returns the
\emph{anchor}. Because the operands come from the graph rather than the
current image, a comparative ranges over every instance the robot has ever
seen. \texttt{between} is the exception in what it returns, not an object but
a passage: the accessible midpoint of two handles. \texttt{expand\_vocab} and
\texttt{verify\_present} maintain the vocabulary the perception process
searches for, and through it what the object graph comes to hold. That graph
lives in the server: following \textsc{SysNav}~\mbox{\cite{sysnav}}, each
LiDAR sweep and the objects recognised in it are folded into persistent
records keyed by identity, so an object seen again from a new viewpoint
updates one record rather than spawning a second, and its extent tightens as
occluded faces come into view. It is where every handle resolves, and where
object reference reads the 3D box it answers with.}

\subsection{Motion}

\rdelar{These 3 tools are the only ones that move the vehicle.}

\rdelar{\texttt{drive\_to} takes a handle, and is where the policy stops owning
actuation. Below it sits a controller with an objective of its own, assumed
only to treat a waypoint as a request rather than a command and not to report
arrival --- neither unusual on a delivered platform. A policy that can neither
command nor read its controller must probe or predict, and probing costs a
drive apiece, so the server predicts: it reimplements the controller's
objective against the same map, finds the point whose predicted \emph{resting
place} lies nearest the handle's coordinate, and publishes that point instead
of the one it wants. Porting to another platform replaces this model and
nothing above it. Arrival is a distance test on odometry, because the
judgement a model makes worst is calling a target reached from a viewpoint
where it is merely visible.}

\rdelar{\texttt{explore} takes no handle and commits to no destination; it exists so a
client holding nothing can still change what the robot can see. When a phrase
will not resolve several attempts running, the fixed goal is dropped and the
robot drives toward the frontier of unseen space; fresh detections pull it
back to reaching, and a coverage plateau ends a fruitless sweep. Where to go
is decided by what the graph lacks, not by a plan the client wrote.}

\rdelar{\texttt{orbit} walks an arc of viewpoints around a handle already reached. A
box measured from one approach is one-sided --- far faces occluded, size
under-read, orientation a guess --- and since every sweep folds into the same
record, the extent tightens as hidden faces rotate into view. \texttt{explore}
grows the set of objects the graph knows; \texttt{orbit} completes the
geometry of one it holds.}

\raddar{In this part we describe motion: the tools that move the vehicle.
\texttt{drive\_to} takes a handle, and is where the policy stops owning
actuation. Below it sits a controller with an objective of its own, assumed
only to treat a waypoint as a request rather than a command and not to report
arrival. A policy that can neither command nor read its controller must probe
or predict, and probing costs a drive apiece, so the server predicts: it
reimplements the controller's objective against the same map, finds the point
whose predicted \emph{resting place} lies nearest the handle's coordinate, and
publishes that point instead of the one it wants. Porting to another platform
replaces this model and nothing above it. Arrival is a distance test on
odometry, because the judgement a model makes worst is calling a target
reached from a viewpoint where it is merely visible.}

\raddar{\texttt{explore} takes no handle and commits to no destination, so a client
holding nothing can still change what the robot can see. When a phrase will
not resolve several attempts running, the fixed goal is dropped and the robot
drives toward the frontier of unseen space; fresh detections pull it back to
reaching, and a coverage plateau ends a fruitless sweep. Where to go is
decided by what the graph lacks, not by a plan the client wrote.
\texttt{orbit} walks an arc of viewpoints around a handle already reached. A
box measured from one approach is one-sided, far faces occluded, size
under-read, orientation a guess, and since every sweep folds into the same
record, the extent tightens as hidden faces rotate into view. \texttt{explore}
grows the set of objects the graph knows; \texttt{orbit} completes the
geometry of one it holds.}

\section{Experiments}
\label{sec:experiments}

%
%
\subsection{Experimental Setup}

\rdelleo{We evaluate on the development set released for the CMU VLN Challenge 2026:
\nScenes{} indoor scenes - single rooms, apartments, offices, a studio -
each paired with two natural-language navigation instructions and, for each
instruction, a ground-truth trajectory demonstrating a correct execution. It
supplies the three things this study needs: the scenes the system drives in,
the sentences it has to follow, and the reference paths every measurement here
is taken against. Every arm in Section~\ref{sec:exp-if} is driven on all
\nQuestions{} questions of all \nScenes{} scenes, so no result rests on a
subset of the benchmark.}

\rdelleo{Two experiments follow, on two corpora. Section~\ref{sec:exp-if} ablates the
system over the \nQuestions{} instruction-following questions, scored on the
trajectory. Section~\ref{sec:results} evaluates the object-reference loop over
a frozen set of \ttQuestions{} comparative referring questions, scored on the
published box; its protocol is stated there because it shares nothing with this
one but the scenes.}

\rdelleo{The instructions have a fixed shape. Some name two destinations in order
go near the stool under the picture and stop at the small table
farthest from the columns and some mix destinations with path constraints
first go to the potted plant furthest from the hookah, then take the
path between the two columns, and stop at the tray on the table. Almost every
destination is fixed by a relative clause rather than a bare noun -the potted plant furthest from the hookah so grounding a phrase
means comparing candidate objects against an anchor, not recognising a
category. That property is what makes the corpus useful here: it exercises
geometric reasoning rather than object recognition. The robot starts from a
fixed pose with no map and has ten minutes to complete an instruction.}

\rdelleo{We implement the scoring protocol ourselves from the semantics of the
instructions and validate it against the ground-truth trajectories. Six points
are split evenly across destinations and constraints. A constraint counts as
met when the driven trajectory passes within 
of the named object;
reaching it out of order keeps half its points; entering a forbidden corridor
subtracts one constraint's worth; the total floors at zero.}

\raddleo{We benchmark on the CMU VLN Challenge 2026 development set, using two of the challenge's task types to exercise different parts of the server. Instruction following
(Section~\ref{sec:exp-if}) drives all \nQuestions{} questions and is scored on
the trajectory; object reference (Section~\ref{sec:results}) answers
\ttQuestions{} comparative referring questions and is scored on a published 3D
box. The two share only the scenes. The reasoning client is Claude Opus~5
(\texttt{claude-opus-5}) with extended thinking disabled and decoding left at the provider default (temperature unset). At the frequency of 2Hz the platform exposes: \SI{1920}{px}$\times$\SI{640}{px} equirectangular ($360^\circ$) RGB image, registered LiDAR,  terrain maps, odometry. The scene graph is built by an open-vocabulary detector cascaded with a segmenter and lifted against LiDAR, running at \SI{\percHz}{\hertz} throughout the run. YOLOv8x-World~v2
(\texttt{yolov8x-worldv2.pt}) detects on a four-face gnomonic projection of the equirectangular panorama; SAM~2.1 Hiera-Large (\texttt{sam2.1\_hiera\_large.pt}) refines each detection to a mask; each mask is back-projected and lifted to a metric box from the registered LiDAR returns it covers, keeping a cluster only when at least \detInliers{} on-surface returns support it. A novel-viewpoint gate (\SI{\detNovelView}{\meter}) suppresses redundant re-detection from nearby poses.
. The detector keeps a
box above a confidence floor of \detScore{}, applies non-maximum suppression at IoU~\detNMS{}, and admits the mask only if SAM's predicted mask-IoU clears \detSAM{}. The detector starts each question from a fixed
prior of \ttPriorSize{} common indoor classes.}

\raddleo{The development set supplies \nScenes{}
indoor scenes-single rooms, apartments, offices, a studio-each paired with
two natural-language navigation instructions and, per instruction, a ground-truth
trajectory. The robot starts from a fixed pose with no prior map. Every question
has a wall-clock budget of \SI{\qBudgetMin}{\minute}, of which the last
\SI{\answerReserveSec}{\second} are reserved for committing the answer, so
navigation is cut off at \SI{\navCutoffSec}{\second}. Each task is evaluated on its own instrument:

\smallskip
\noindent\emph{Instruction following} (driven trajectory):
\begin{itemize}\itemsep1pt \parskip0pt \topsep1pt
  \item {Challenge Score} (0--6) $\uparrow$: 6 points split evenly across destinations and path constraints (met within 1.25\,m; half credit for out-of-order arrival; deduction for entering keep-out corridors; floored at zero).
  \item {nDTW} $\uparrow$: normalised dynamic-time-warping agreement with the ground-truth reference trajectory.
 \item {TL ratio}: ratio of driven trajectory length to ground truth (1.0 is optimal).
  \item {Closest approach} (m) $\downarrow$: minimum distance reached to each referenced destination.
\end{itemize}

\smallskip
\noindent\emph{Object reference} (published oriented box):
\begin{itemize}\itemsep1pt \parskip0pt \topsep1pt
  \item {Hit rate} (IoU $\ge \ttIoUThresh{}$) $\uparrow$: fraction of published boxes clearing the 3D overlap threshold.
  \item {Mean IoU} $\uparrow$: average 3D bounding box overlap with the ground-truth box.
  \item {Centre distance} (m) $\downarrow$: threshold-free Euclidean distance from the published marker to the ground-truth centre.
\end{itemize}}

\subsection{Instruction Following}
\label{sec:exp-if}


\begin{table}[t]
  \centering
  \caption{Removing one component at a time. \nQuestions{} questions on
  \nScenes{} scenes. $\Delta$ (points of score) and $t$ are paired against the
  full system, repeats averaged first, and $|t| > 2.05$ resolves at
  $\mathrm{df} = 29$. TL is trajectory length as a ratio to the ground truth,
  which averages \tlrRef{}. Approach is the closest approach to the referenced
  destination (Section~\ref{sec:exp-setup}), over hits and misses alike; its
  own paired tests are in the text. Two further arms are in
  Appendix~\ref{app:arms}.}
  \label{tab:arms}
  \small
  \setlength{\tabcolsep}{3pt}
  \begin{tabular}{@{}lcccccc@{}}
    \toprule
    Arm & Score & $\Delta$ & $t$ & nDTW & TL & Approach \\
    \midrule
    Full system        & \armOurs{} & -         & -                 & \ndtwOurs{} & \tlrOurs{} & \apprOurs{} \\
    w/o Lift           & \armLift{} & $-\dLift$ & $\tLift$          & \ndtwLift{} & \tlrLift{} & \apprLift{} \\
    w/o Platform Model & \armPlat{} & $-\dPlat$ & $\mathbf{\tPlat}$ & \ndtwPlat{} & \tlrPlat{} & \apprPlat{} \\
    \bottomrule
  \end{tabular}
\end{table}

\raddleo{Only the platform model resolves on the score. Removing it costs
\dPlat{} points at $t = \tPlat$; removing the lift costs \dLift{}, a twelfth of the
\minDetect{}-point difference this design can resolve (Appendix~\ref{app:power}),
so its point estimate - not merely the interval around it - is consistent with
zero. This is not because the lift is inaccurate: its scan range has a median
absolute error of \estLift{}, against \estModel{} for the model's own
\texttt{distance\_m}, which is worse than always answering \SI{5.0}{\meter}. The
platform simply absorbs that accuracy - the waypoint node re-plans every
published point against its own terrain map, so a median \aimGapMed{} difference
in aim becomes only \settleGapMed{} in where the vehicle stops - so the lift's
real return is cost, not score: \dSteps{} fewer grounding calls per question
($t = \tSteps$).}

\raddleo{What the score cannot see, distance can. The two ablations fail
in opposite directions: \emph{w/o Lift} drives $\tlrLift{}\times$ the reference
because the model's metres overshoot, \emph{w/o Platform Model} $\tlrPlat{}\times$
because an unmodelled free-space check only ever pulls a waypoint in; removing
both drives $\tlrNaive{}\times$, indistinguishable from correct, yet scores
\dNaive{} below the full system at $t = \tNaive$ - driving the right
\emph{distance} is not driving to the right \emph{place}. Closest approach makes
this legible (Table~\ref{tab:arms}, last column): removing the platform
model leaves the robot \dApprPlat{} further away on \signApprPlat{} questions at $t = \tApprPlat$,
against $t = \tPlat$ on the score, and since the arrival tolerance is \tol{} that
gap is much of the margin between arriving and not. The two instruments agree in
direction but the metre one resolves earlier - on the \nQuestionsOld{} questions
available before the last two scenes it had already separated both the platform
model and the both-removed arm from the full system, four questions ahead of the
score. A benchmark scored on arrival within a fixed radius needs more of itself to
say the same thing.}

\subsection{Object Reference}
\label{sec:results}

\raddleo{End-to-end, the full loop hits \ttFullPct{} of the \ttQuestions{}
referring questions (mean IoU \ttFullIoU{}, Table~\ref{tab:objref}), and on a hit
the box centre lands a mean \SI{36}{\centi\meter} from truth - modest coverage,
but geometrically tight when it commits. Two ablations each remove one half of
the rule \emph{``the VLM proposes semantics; geometry decides metrics''} to
isolate where that precision comes from.}

\raddleo{\begin{table}[t]
  \centering
  \caption{End-to-end object reference across all \ttScenes{} scenes
  (\ttQuestions{} paired questions; a hit is a 3D IoU $\ge \ttIoUThresh{}$ with
  the ground-truth box). \emph{Naive} strips the MCP tool structure -
  navigation and the answer become free text; \emph{model-metric} keeps
  perception for navigation but lets the model emit the answer's $(x,y,z)$
  itself, bypassing the scene graph at answer time.}
  \label{tab:objref}
  \begin{tabular}{lcc}
    \toprule
    Arm & Hit (IoU$\,\ge\ttIoUThresh{}$) & Mean IoU \\
    \midrule
    Full system (grounded) & \textbf{\ttFullPct} & \textbf{\ttFullIoU} \\
    \;\;w/o tool structure (naive)       & \ttNaivePct  & \ttNaiveIoU \\
    \;\;w/o geometric grounding (metric) & \ttModelPct  & \ttModelIoU \\
    \bottomrule
  \end{tabular}
\end{table}}

\begin{figure}[t]
  \centering
  \includegraphics[width=\columnwidth]{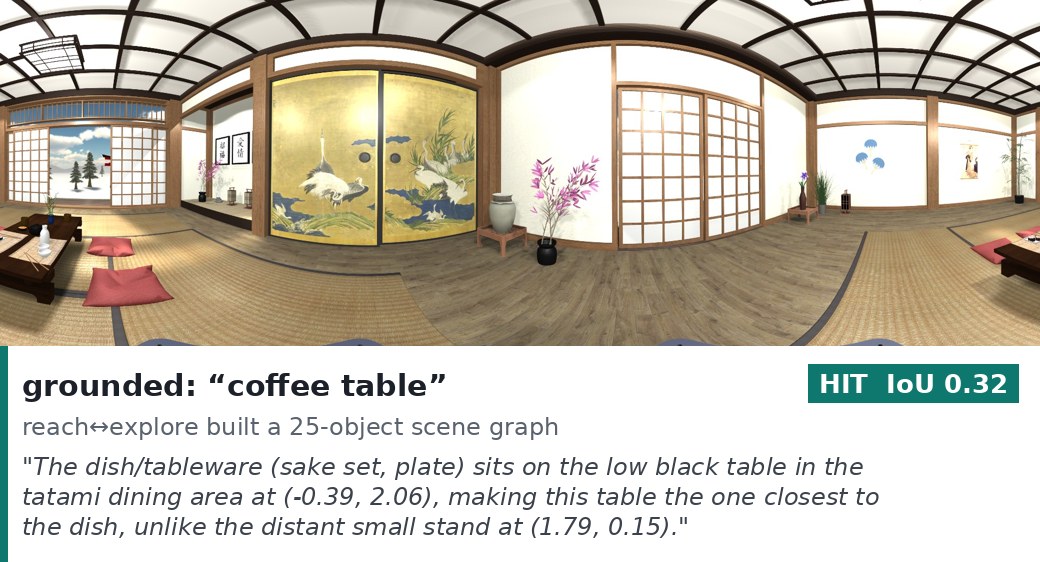}\\[3pt]
  \includegraphics[width=\columnwidth]{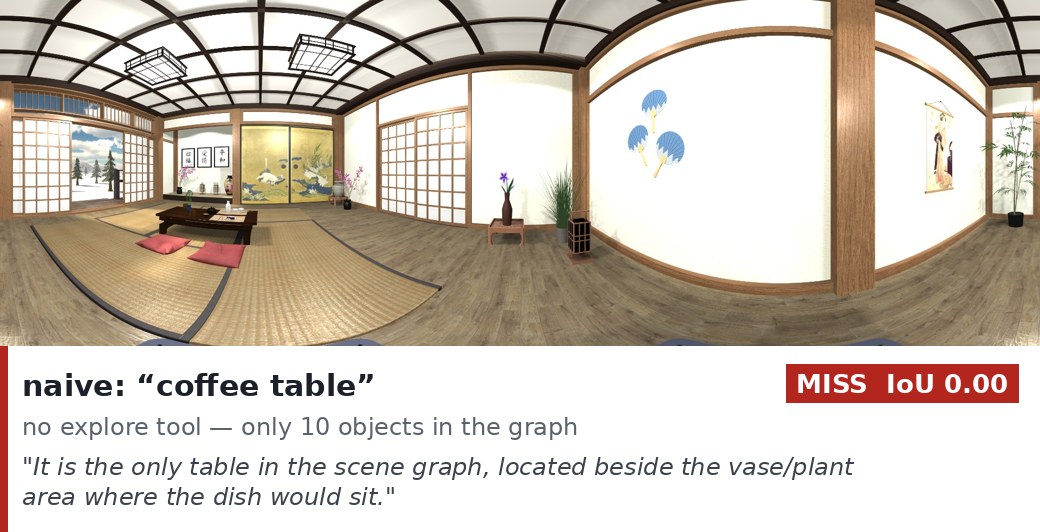}
  \caption{\textbf{Exploration robustness.} Same question
  (\emph{``the table closest to the dish''}). The reach$\leftrightarrow$explore
  controller builds a \ttCaseFullObjs{}-object scene graph and picks the low
  tatami table beside the dish; the naive arm, with no \texttt{explore} tool,
  gathers only \ttCaseNaiveObjs{} objects and picks ``the only table in the scene
  graph,'' the wrong one.}
  \label{fig:explore}
\end{figure}

\begin{figure}[t]
  \centering
  \includegraphics[width=\columnwidth]{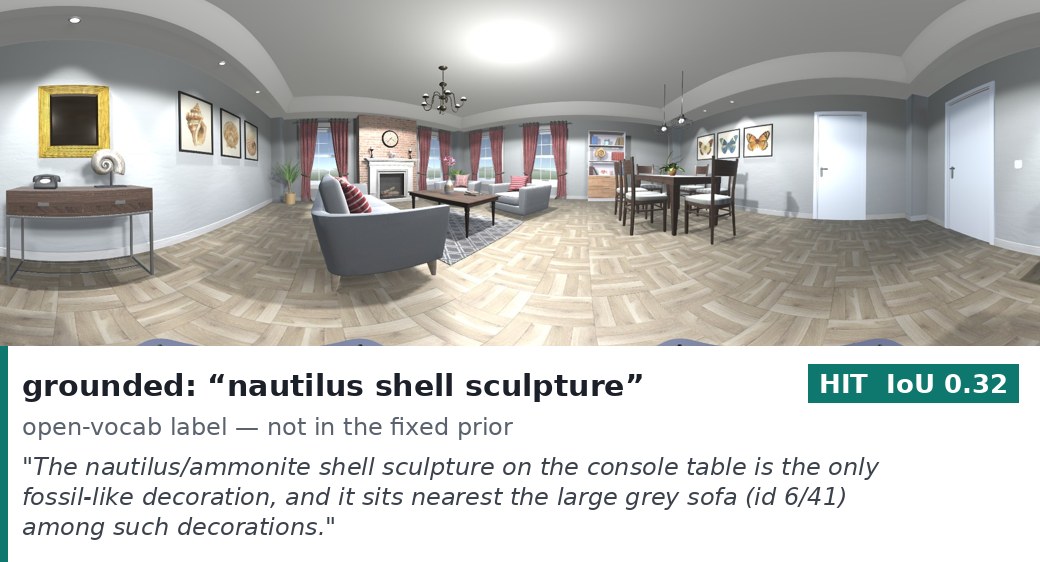}\\[3pt]
  \includegraphics[width=\columnwidth]{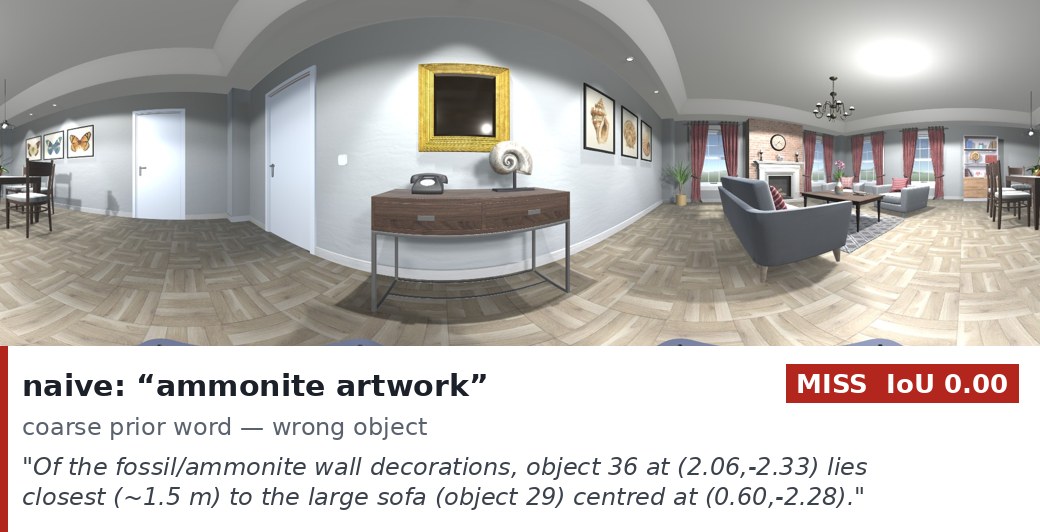}
  \caption{\textbf{Vocabulary injection.} Same question
  (\emph{``the fossil decoration closest to the big sofa''}). Because
  \texttt{expand\_vocabulary} primes the detector from the question, the full
  system labels the console-table ornament \emph{nautilus shell sculpture} - a
  phrase absent from the fixed \ttPriorSize{}-class prior - and grounds it; the
  naive arm falls back to the nearest prior word, \emph{ammonite artwork}, and
  commits to a wall picture.}
  \label{fig:vocab}
\end{figure}

\raddleo{The tool schema is modest; the tools it wraps are not. The \emph{naive} arm removes the MCP tool layer entirely: the model navigates and
answers in free text, with no \texttt{explore}, \texttt{expand\_vocabulary} or
\texttt{verify\_present} to shape the scene graph it draws its answer from. Removing the MCP structure (\emph{naive}) barely moves the metric - \ttFullPct{}
$\rightarrow$ \ttNaivePct{}, mean IoU essentially unchanged - so the interface
is not itself load-bearing. What it strips are \texttt{explore},
\texttt{expand\_vocabulary} and \texttt{verify\_present}, and their loss shows in
kind rather than in score. Without \texttt{explore} the naive arm reasons over a
starved graph - the full system's is larger on \ttGraphBigger{} of
\ttQuestions{} questions (mean \ttGraphBaseline{} vs.\ \ttGraphNaive{} objects,
Figure~\ref{fig:explore}) - and without vocabulary priming it falls back to a
coarser word (\emph{nautilus shell sculpture} $\rightarrow$ \emph{ammonite
artwork}, Figure~\ref{fig:vocab}), so it picks the right label from the wrong
set.}

\raddleo{When the model emits the
answer's $(x,y,z)$ itself (\emph{model-metric}), bypassing the scene graph at
answer time, not one of the \ttQuestions{} answers clears the threshold
(\ttModelHits{}/\ttQuestions{}, mean IoU \ttModelIoU{}): the arm reports no
extent - a default cube - leaving nothing for an overlap metric to credit.
The failure is geometric, not semantic. Asked for \emph{``the bedroom light above
the bed closest to the towel,''} the grounded system reasons over coordinates and
lands \ttCaseBaseDist{} from truth, while this arm merely estimates ``about
\ttCaseMetricEst{} in that direction'' and misses by \ttCaseMetricDist{}. Same
model, same navigation, same scene - only the source of the final box differs.}

\subsection{Qualitative Analysis}
\label{sec:exp-traces}

\begin{figure}[t]
  \centering
  \includegraphics[width=\columnwidth]{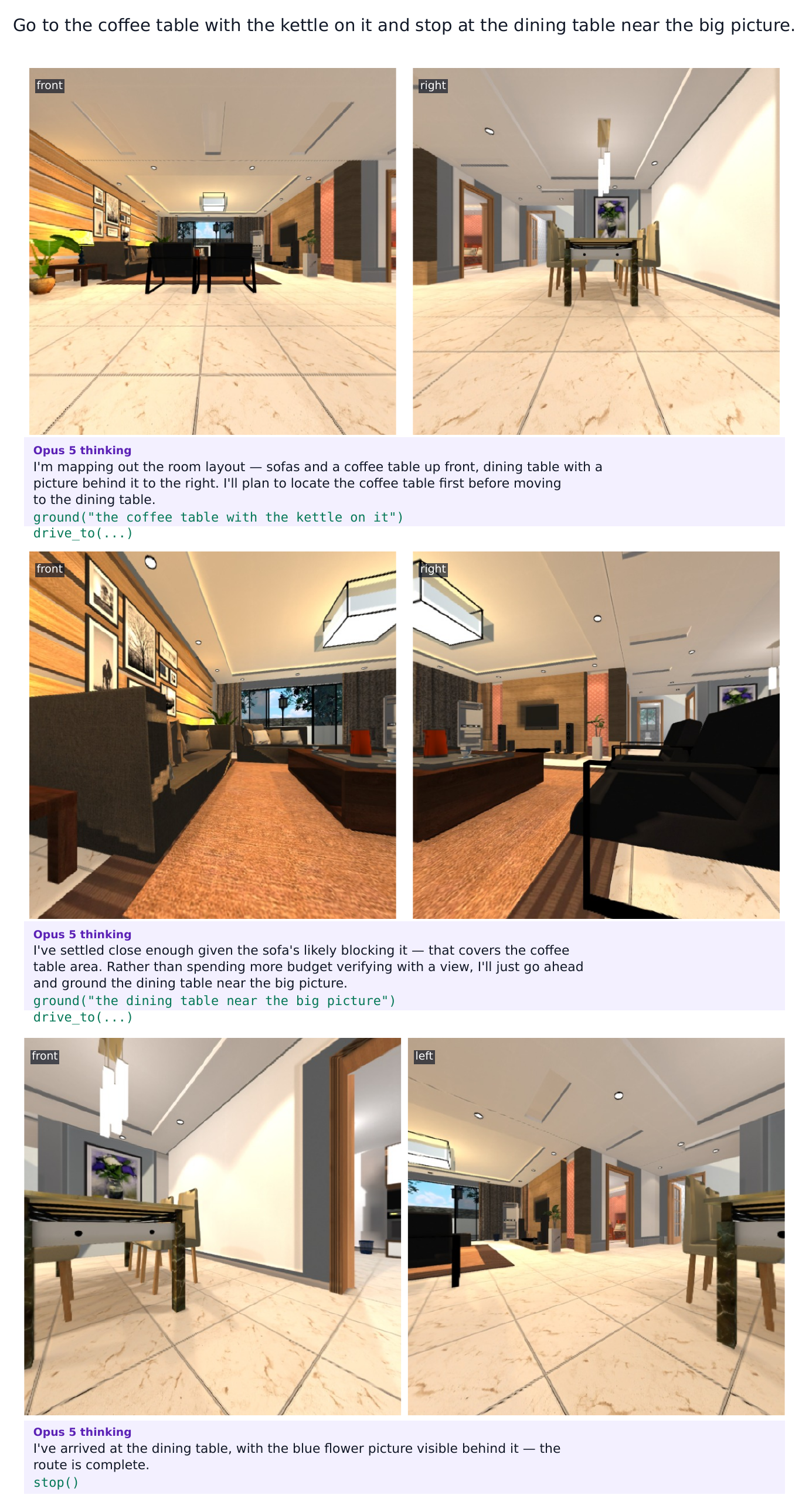}
  \caption{Tool-call trace on a Type-3 instruction-following episode
  (\emph{``Go to the coffee table with the kettle on it and stop at the
  dining table near the big picture.''}). Each row is one decision:
  the two faces the client received, its reasoning, and the tools it
  issued. The client grounds and drives to the coffee table from spawn,
  skips a verification view after settling beside the sofas, then
  grounds and drives to the dining table and calls \texttt{stop}.}
  \label{fig:qual-trace}
\end{figure}

In this section we qualitatively analyse the tool-call trace on a
Type-3 instruction-following episode (Fig.~\ref{fig:qual-trace}). The
instruction imposes two ordered goals: first the coffee table with the
kettle, then the dining table near the big picture. In the first image
pair, captured at spawn, the model reads the living room ahead and the
dining alcove to the right, grounds the coffee-table constraint, and
immediately issues \texttt{drive\_to}. After settling beside the
sofas-close enough that the table is partly occluded-it skips a
verification view, grounds the dining-table constraint from that new
viewpoint, and drives again. The last pair is taken at the dining table
with the floral picture in view; the model then calls \texttt{stop},
completing both clauses in order.

\section{Discussion and Limitations}
\label{sec:lessons}
We have both qualitative and quantitative results from running the system
with the challenge as a testbed, and together they suggest that this is a
promising way to navigate without a prior map. Several limitations bound
what we can claim from them. First, latency is high: a model call and the
step it triggers cost roughly half a minute, and this cost is structural
rather than incidental, since every tool the client invokes waits on a
round trip to a frontier model; deployment outside a simulator would
require a faster model in the \texttt{ground} role, a client that commits
to several steps per call, or both. Second, every run uses a single
frontier model, and the interface was designed against that model's
failure modes, so whether a handle-typed boundary matters more for a
weaker model remains untested. Third, our evaluation covers one
simulator, one robot, and no real hardware; the controller model inside
\texttt{drive\_to} is the component most tied to this deployment, and
while porting replaces it and nothing above it, that is a claim we make
rather than one we have run. Finally, the official evaluator is closed,
so the arrival tolerance and the instance-matching rule are both ours
and both move the result; we report the pair rather than the flattering
member of it, and every number is from the development set.
\section{Conclusion}
\label{sec:conclusion}
\textsc{Embodied-Nav-MCP} cuts an open-vocabulary navigation stack into 10
tools a VLM agent drives at inference time, typed so that every argument is a
phrase or a handle and no tool accepts a number. On the CMU Vision-Language
Navigation Challenge 2026 the full configuration scores \armOurs{} across all
\nQuestions{} instruction-following questions on \nScenes{} scenes and clears
the object-reference overlap threshold on \ttFullHits{} of \ttQuestions{}
questions, while the configuration that predates the rule --- metric
quantities taken from the model, no model of the controller --- scores
metric estimate is worth only what the actuation interface preserves: the same
scan lift that decides object reference outright is almost invisible in
navigation, because the platform re-plans whatever it is handed. Measuring
better is worth doing. Measuring better without knowing what the stack beneath
does with the measurement is not.

\bibliographystyle{IEEEtran}
\bibliography{refs}

%

\newpage
\appendix
%
%
%
\section{Appendix: Resolution, and the Two Arms Held Out of
Section~\ref{sec:experiments}}
\label{app:arms}

Five arms were driven; Table~\ref{tab:arms} reports three. This appendix gives
what the design can resolve, and then the two omitted arms in full, with the
analyses that failed as well as the ones that did not --- an arm reported only
when its result is convenient is not an ablation. Neither arm is held back for
being weak: one of the two is resolved.

\begin{table}[h]
  \centering
  \caption{All five arms over the same \nQuestions{} questions. $\Delta$ and
  $t$ are paired against the full system. Two arms clear $|t| > 2.05$, and both
  are arms that stop predicting where the waypoint will settle.}
  \label{tab:allarms}
  \small
  \begin{tabular}{@{}lccccc@{}}
    \toprule
    Arm & Score & $\Delta$ (pp) & $t$ & nDTW & TL ratio \\
    \midrule
    Full system        & \armOurs{}  & ---        & ---   & \ndtwOurs{}  & \tlrOurs{}  \\
    w/o Lift           & \armLift{}  & $-\dLift$  & $\tLift$ & \ndtwLift{}  & \tlrLift{}  \\
    w/o Platform Model & \armPlat{}  & $-\dPlat$  & $\mathbf{\tPlat}$ & \ndtwPlat{} & \tlrPlat{} \\
    \midrule
    Naive (neither)    & \armNaive{} & $-\dNaive$ & $\mathbf{\tNaive}$ & \ndtwNaive{} & \tlrNaive{} \\
    w/o Memory         & \armMem{}   & $+\dMem$   & $\tMem$   & \ndtwMem{}   & \tlrMem{}   \\
    \bottomrule
  \end{tabular}
\end{table}

\subsection{What this design can resolve}
\label{app:power}

Two configurations were driven more than once, which makes the spread between
repeats of an \emph{identical} configuration measurable rather than assumed.
The full system, re-driven, scores \repeatOurs{}; \emph{w/o Lift}'s two passes
score \repeatLift{}. The larger of those gaps is \repeatWorst{} points, from
changing nothing at all. A single run's per-question score has a standard
deviation of \sdWithin{}/6 about that question's own mean against
\sdBetween{}/6 between questions, so roughly a quarter of the variance in a
corpus mean is run-to-run noise rather than question difficulty. At
\nQuestions{} questions that puts the smallest resolvable difference between
two single-pass arms near \minDetect{} points. \emph{Naive} clears that floor
and \emph{w/o Memory} does not, which is what separates the two subsections
below: the first reports a result, the second an observation.

\subsection{Removing both components}

\emph{Naive} takes the range from the model and does not predict the
controller. It is the configuration a team arrives at before reading the
controller's source, and it exists to test whether the two effects compound.
They do not. Removing the lift costs \dLift{} points and removing the platform
model costs \dPlat{}; an additive account predicts \SI{50.3}{\percent} and the
arm scores \armNaive{}, above \emph{w/o Platform Model}'s \armPlat{}. That
the arm with both faults outscores the arm with one is the observation; the
gap between them, $+3.3$ points at $t = 0.83$, is not resolved and we claim
nothing from it.

What is resolved is the arm itself: \dNaive{} points below the full system at
$t = \tNaive$, on \num{8} of the \num{9} questions where the two differ. It
was not resolved on the \nQuestionsOld{} questions available before the last
two scenes were unpacked ($8.3$ points at $t = 1.88$), and the four added
questions carried it over rather than changed it --- the arm scores
\SI{54.2}{\percent} on those four against \SI{54.5}{\percent} on the other
\nQuestionsOld{}. The conclusion it supports is the main text's: what costs
marks is not knowing where the waypoint will settle, and it costs them whether
or not the range that reached the waypoint was measured.

The mechanism the observation suggests is visible in the geometry and does not
survive a test. Path length behaves exactly as error cancellation predicts:
\emph{w/o Platform Model} drives $\tlrPlat{}\times$ the reference length, and
adding the model's systematically long range brings that to $\tlrNaive{}\times$
($t = 2.61$, resolved). But the per-question score gain does not track the
per-question shortfall ($r = 0.27$, $t = 1.45$), and splitting the corpus at
the median shortfall gives the same gain in both halves. Correcting the
distance is real; that the correction is what earns the marks is not
established, and we do not claim it.

In metres \emph{Naive} is \dApprNaive{} further from the referenced
destination than the full system ($t = \tApprNaive$), also resolved. This is
the comparison Section~\ref{sec:exp-metric} refers to: on the
\nQuestionsOld{} questions driven first it was clear in metres while the score
could not see it, and the score only caught up once the last two scenes were
added.

\subsection{Removing the prompt's memory}

The shipped prompt carries a block listing the poses the robot has already
occupied, with an instruction to prefer somewhere it has not stood. Removing
it is a prompt-only change: the drive loop's own refusal to re-visit a spent
bearing runs off separate state and is untouched.

On the four quantities the block's own wording predicts, removing it changes
nothing measurable. Revisit rate is \SI{27}{\percent} against
\SI{28}{\percent}, positional spread \SI{2.64}{\meter} against
\SI{2.72}{\meter}, steps per question $8.9$ against $8.8$, and the share of
steps from which the target is visible \SI{94}{\percent} against
\SI{94}{\percent}, with paired $t$ of $-0.10$, $-0.64$, $0.11$ and $0.12$
and every sign test near-tied. These are per-step measurements and carry
roughly ten times the samples a per-question score does.

The score moves the other way: \dMem{} points \emph{in favour} of removing
the block, at $t = \tMem$. That gap is smaller than the \repeatWorst{}-point
spread we measure between two passes of an identical configuration
(Appendix~\ref{app:power}), and the ablation perturbs fewer questions
(\num{6} of \nQuestions{}) than a plain re-drive does (\numrange{8}{9}). Six
questions differ at all: three are passage constraints flipping, one for the
arm and two against; one is a \SI{17}{\centi\meter} near-miss past the
\tol{} tolerance; one is both runs exhausting the step budget. The remaining
one has a legible mechanism that does not reproduce on the other two, and we
report it as an anecdote rather than a finding.

The defensible statement is therefore narrow and negative: removing the block
does not hurt. The organisers' own reference trajectories double back over
\SI{15}{\percent} of their length on average, and \num{21} of the \num{30}
do it at all --- returning within \SI{0.5}{\meter} of a point they left at
least a metre of travel earlier --- so an instruction to prefer unvisited
ground argues against what a correct execution does.

\end{document}